\documentclass[letterpaper, 10 pt, conference]{ieeeconf}  

\IEEEoverridecommandlockouts                              

\usepackage{graphics} 
\usepackage{epsfig} 
\usepackage{mathptmx} 
\usepackage{times} 
\usepackage{amsmath} 
\usepackage{amssymb}  
\usepackage{geometry}
\usepackage{color}
\usepackage{algorithm,algpseudocode}
\usepackage{babel}
\usepackage{multirow}
\usepackage{amstext}
\usepackage{units}
\usepackage{booktabs}
\usepackage{caption}
\usepackage{subcaption}
\usepackage{xcolor}
\newcommand{\gray}[1]{\textcolor{gray}{#1}}

\title{\LARGE \bf
Multi-Class 3D Object Detection with Single-Class Supervision
}

\author{Mao Ye$^\mathsection$, Chenxi Liu$^{\mathparagraph \dagger}$, Maoqing Yao$^\mathparagraph$, Weiyue Wang$^\mathparagraph$, Zhaoqi Leng$^\mathparagraph$, Charles R. Qi$^\mathparagraph$, Dragomir Anguelov$^\mathparagraph$
\thanks{$^\mathsection$The University of Texas at Austin \quad $^\mathparagraph$Waymo \quad $^\dagger$Corresponding author}
}

\begin{document}

\global\long\def\D{\mathcal{D}}%
\global\long\def\L{\text{L}}%
\global\long\def\th{\boldsymbol{\theta}}%
\global\long\def\seg{\text{seg}}%
\global\long\def\hm{\text{hm}}%
\global\long\def\bbox{\text{bbox}}%
\global\long\def\ri{\text{ri}}%
\global\long\def\fm{\text{fm}}%
\global\long\def\1{\boldsymbol{1}}%
\global\long\def\up{\text{U}_{\text{pixel}}}%
\global\long\def\cp{\text{C}_{\text{pixel}}}%
\global\long\def\po{\text{point}}%
\global\long\def\uh{\text{U}_{\text{hm}}}%
\global\long\def\pse{\text{pseudo}}%
\global\long\def\mar{\text{marginal}}%
\global\long\def\teach{\text{teacher}}%
\global\long\def\score{\text{score}}%
\global\long\def\p{\text{p}}%
\global\long\def\y{\text{y}}%
\global\long\def\h{\text{h}}%
\global\long\def\X{\text{X}}%
\global\long\def\Y{\text{Y}}%
\global\long\def\uc{\text{U}_{\text{class}}}%

\newcommand{\todo}[1]{\textcolor{red}{TODO: #1}}

\maketitle
\thispagestyle{empty}
\pagestyle{empty}

\begin{abstract}


While multi-class 3D detectors are needed in many robotics applications, training them with fully labeled datasets can be expensive in labeling cost. An alternative approach is to have targeted single-class labels on disjoint data samples. In this paper, we are interested in training a \emph{multi}-class 3D object detection model, while using these \emph{single}-class labeled data. We begin by detailing the unique stance of our ``Single-Class Supervision'' (SCS) setting with respect to related concepts such as partial supervision and semi supervision. Then, based on the case study of training the multi-class version of Range Sparse Net (RSN), we adapt a spectrum of algorithms --- from supervised learning to pseudo-labeling --- to fully exploit the properties of our SCS setting, and perform extensive ablation studies to identify the most effective algorithm and practice. Empirical experiments on the Waymo Open Dataset show that proper training under SCS can approach or match full supervision training while saving labeling costs.
\end{abstract}

\section{Introduction}
\begin{figure}[t]
     \centering
     \begin{subfigure}[b]{1.\columnwidth}
         \centering
         \includegraphics[scale=0.44]{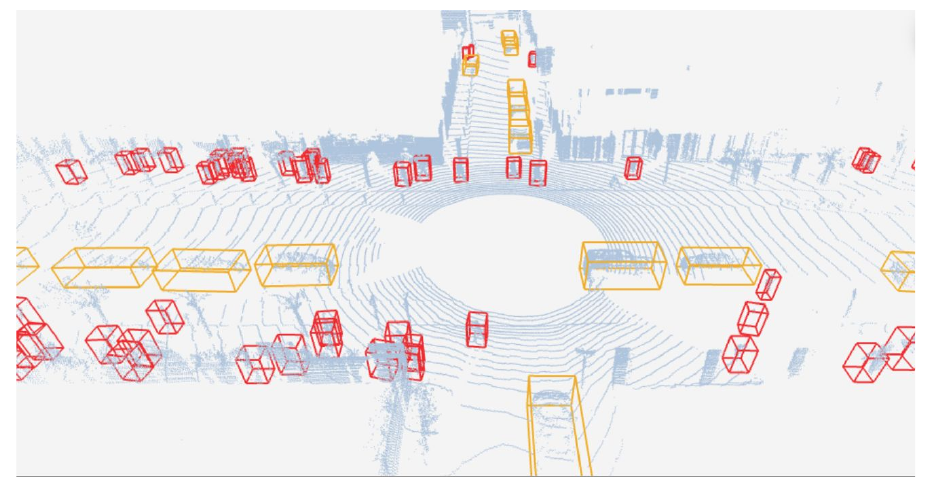} 
         \includegraphics[scale=0.44]{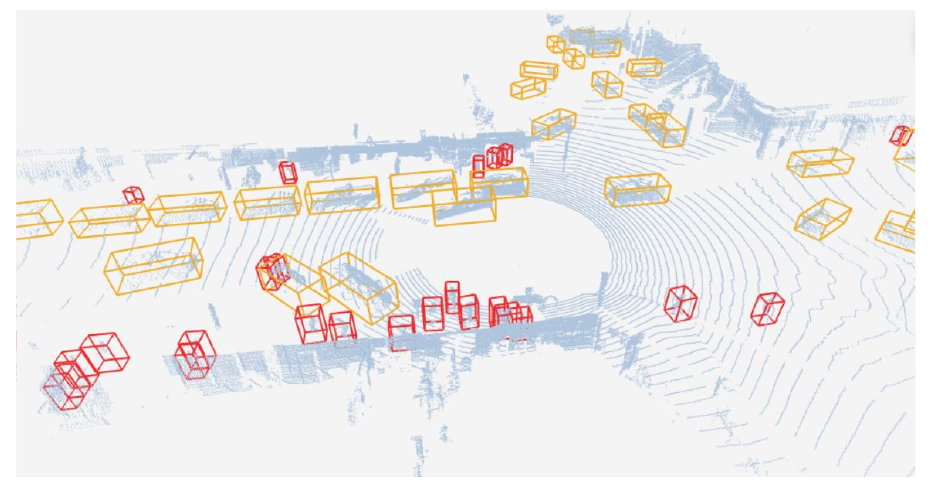} 
         \caption{Full Supervision}
         \label{fig: scs_full}
     \end{subfigure}
     \begin{subfigure}[b]{1.\columnwidth}
         \centering
         \includegraphics[scale=0.44]{fig/scs_12A.pdf} 
         \includegraphics[scale=0.44]{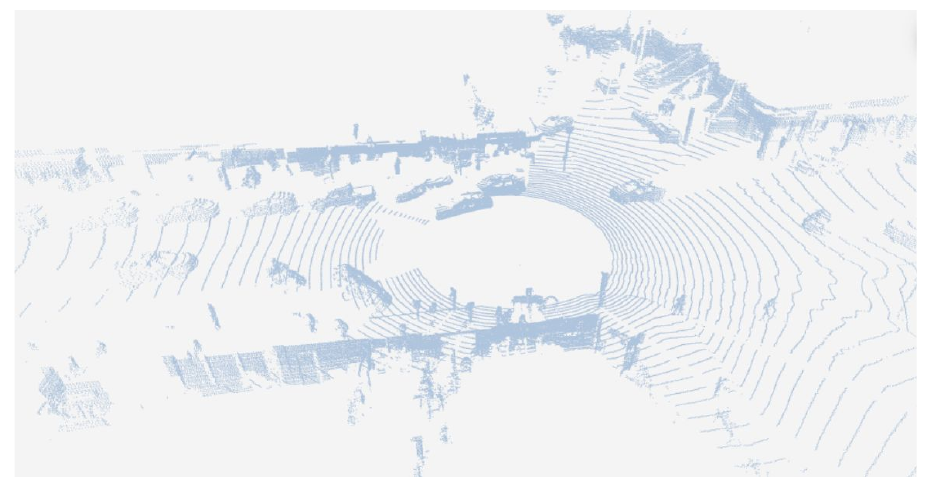} 
         \caption{Semi Supervision}
         \label{fig: scs_semi}
     \end{subfigure}
     \begin{subfigure}[b]{1.\columnwidth}
         \centering
         \includegraphics[scale=0.44]{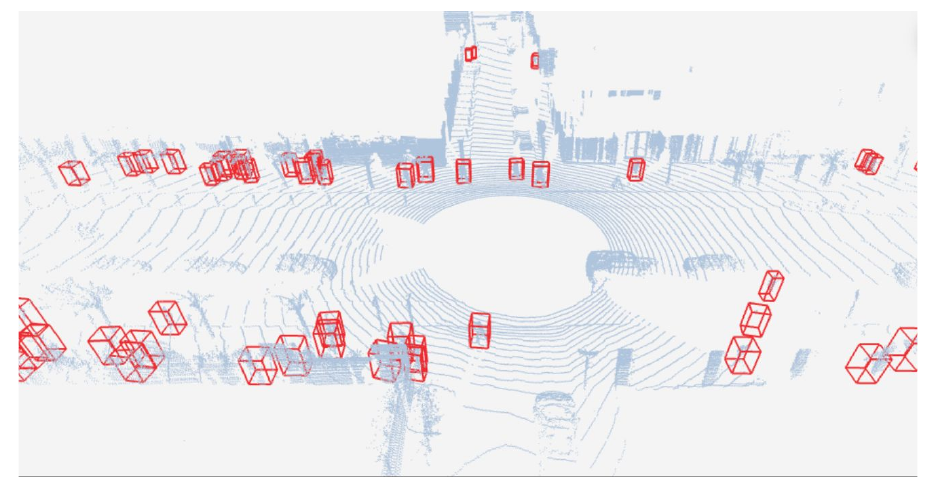} 
         \includegraphics[scale=0.44]{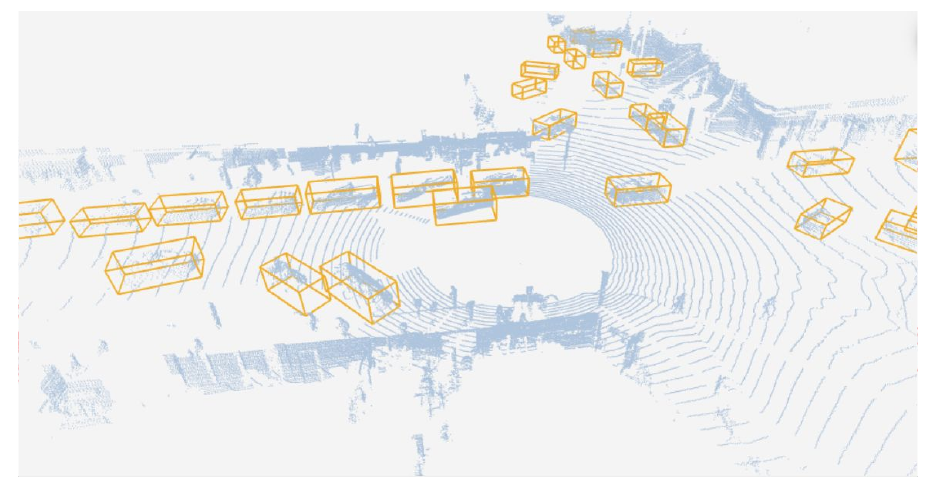} 
         \caption{Single-Class Supervision (ours)}
         \label{fig: scs_ours}
     \end{subfigure}
        \caption{
        Full supervision (a), where all objects are properly labeled, is ideal in training a multi-class detector.
        Semi supervision (b) is a classic setting where each image is either fully labeled or not labeled.
        Single-class supervision (c) is an under-explored setting that we study. There is only one class of objects labeled for each image.
        }
        \label{fig: scs}
\end{figure}

3D object detection is a core component in various robotics and autonomous driving applications.
Existing public datasets \cite{geiger2012we,huang2018apolloscape,behley2019semantickitti,sun2020scalability} often provide the labels for all $K$ classes on all data, which enables what we call full supervision training (Figure~\ref{fig: scs_full}).
However, such fully labeled datasets are not scalable in real-world applications at the industrial scale, given the cost of labeling.
A sensible approach then, is to perform targeted labeling for each class of interest, resulting in $K$ datasets that are possibly non-overlapping.
We term this the ``Single-Class Supervision'' (SCS) setting (Figure~\ref{fig: scs_ours}).

Intuitively the SCS setting enjoys many advantages, such as dedicated allocation of labeling resources (e.g. only label the ``rare / hard'' class and save the cost of labeling the ``common / easy'' class), and better control of class-specific performance metrics (e.g. label more data for a certain class if the model's performance on that class is not accurate enough). Under this setting, it is straightforward to train a single-class detector, which is common in 3D detection for autonomous driving application \cite{zhou2018voxelnet,yan2018second,meyer2019lasernet,lang2019pointpillars,zhou2020end,shi2020pv,sun2021rsn}. However, the training protocol becomes unclear when training a multi-class detector: since every training example has incomplete labels, even conducting supervised learning becomes challenging\footnote{
When objects from different classes tend not to co-exist, SCS is easy as it is essentially fully supervised. But in this paper we target the opposite.
}. 

In this paper, we describe ways to learn under the SCS setting based on the case study of training a multi-class Range Sparse Net (RSN) \cite{sun2021rsn}, which is a state-of-the-art 3D object detection model. 
The RSN model is a two-stage detector. It performs segmentation as its first stage, which makes our solutions potentially generalizable to other tasks such as semantic segmentation as well. 
We find that it is critical to correctly handle the missing label property of SCS, and propose an Informed supervision scheme that makes the most out of the (partial) labels provided.
Based on the Informed supervision scheme, from simple to complex, we adapt supervised learning algorithms (Section~\ref{sec: supervised}) and pseudo-labeling algorithms (Section~\ref{sec: pseudo}) under the SCS setting.
Other important training techniques, including dataset resampling and combining pseudo labels from different sources, are also studied.
On the Waymo Open Dataset \cite{sun2020scalability}, we conduct experiments to demonstrate that proper modeling under SCS can approach or match the full supervision scenario, showing its practicality and promise in saving (re)labeling cost.

\section{Defining\&Situating Single-Class Supervision}
\label{sec: definition}

We begin by considering object detection. 
Let $M$ be the number of images (range images for the 3D case), and $K$ be the number of classes in the dataset.
Each image, indexed by $i$, contains $N_i$ objects of various classes, and $N = \sum_{i=1}^M N_i$ is the total number of objects in the dataset.

\subsection{Full Supervision} Full supervision is the situation where all $N$ objects are labeled.
This is of course ideal, and the model performance under full supervision can be considered the upper bound to any of the partial supervision situations discussed next.

\subsection{Partial Supervision}
We use the term partial supervision to summarize all situations where $n < N$ objects are labeled. While the partiality can be from multiple aspects, there are several important special cases of partial supervision, including semi supervision and (our) single-class supervision.

\subsubsection{Semi Supervision}
Research on semi-supervised learning has mostly been conducted on the image classification setting \cite{grandvalet2005semi,lee2013pseudo,laine2016temporal,sajjadi2016regularization,tarvainen2017mean,miyato2018virtual, lee2013pseudo,yalniz2019billion, berthelot2019remixmatch, xie2020self, sohn2020fixmatch}, with interest in segmentation and object detection only rising recently \cite{french2019semi,zou2020pseudoseg,kim2020structured,ouali2020semi,chen2021semi,zhu2020improving,feng2020semi,rosenberg2005semi,jeong2019consistency,tang2019transferable, tang2021proposal, sohn2020simple, yang2021interactive, liu2021unbiased, zhao2020sess, wei2021semantic, wang20213dioumatch, caine2021pseudo}.
We follow these last set of works and characterize the semi supervision setting as: (upon sorting,) for a certain integer $S$, full labels are provided for images with indices $1 \leq i \leq S$, and no labels for images with indices $S < i \leq M$.

\subsubsection{Single-Class Supervision}

We define (our) single-class supervision (SCS) as: for each image $i$, only objects from $1$ class $C_i \in \{1, \hdots, K\}$ are labeled. 
The collection of labels $C_i$ cover all $K$ classes. 
We illustrate full supervision, semi supervision, and single-class supervision in Figure~\ref{fig: scs} for intuitive comparison.

\subsubsection{Extensions to Single-Class Supervision}

While this paper focuses on the single-class supervision defined above, there are natural extensions / relaxations. 
For example, we can relax the ``only $1$ class'' constraint to having $\geq 1$ classes, or having $\leq 1$ classes. 
The latter implies some images may have no labels, making it closer to semi supervision, to become a hybrid between semi supervision and SCS.

Under the framework of this section, image classification can be viewed as $N_i = 1, \forall i$\footnote{Admittedly there may be more than one object in an image used for classification, but here we make this simplification to illustrate the collapse.}. 
This effectively collapses semi supervision and SCS to the same setting.
\section{Multi-Class Range Sparse Net}

Our multi-class 3D object detector is based on Range Sparse Net (RSN) \cite{sun2021rsn}, which is a state-of-the-art single-class detector. In this section, we recap its architecture and describe how we extend it for the multi-class setting.

\subsection{Architecture}
RSN is a two-stage detector. In the first stage, the model segments LiDAR range image pixels into two categories: background and foreground. In the second stage, points classified as foreground are voxelized and fed into a sparse convolution network followed by a detection head to predict the 3D bounding boxes.

The RSN architectures described in \cite{sun2021rsn} perform single-class 3D object detection.
We extend RSN to multi-class by sharing the first stage among all $K$ classes, but keeping individual second stages for each class. 
This means the foreground point selection stage is now a $K + 1$-way instead of a $1 + 1 = 2$-way segmentation, with index $0$ being the background.
In the second stage, only points that are classified as foreground class $k$ are voxelized and fed into the corresponding layer. 
We choose not to share the second stage, as different object classes may call for different voxelization granularities (e.g. pedestrians require a finer resolution than vehicles). 

\subsection{Losses for training RSN}

Below we describe how we extend the single-class training loss in RSN~\cite{sun2021rsn} to the multi-class setting.
\subsubsection{Foreground Point Selection}
In a training sample $(\X_{\ri}, \Y_{\bbox})$, $\X_{\ri} \in \mathbb{R}^{H \times W \times 3}$ is the LiDAR range image, and $\Y_{\bbox} = \cup_{k=1}^K \Y_{\bbox}^k$ is the union of the labeled 3D bounding boxes of all $K$ classes. 
By examining whether a LiDAR point lies within any of the labeled 3D bounding boxes, we can generate the ground truth for foreground point selection $\Y_{\ri} \in \mathbb{Z}^{H \times W \times K + 1}$ which is a one-hot vector for each pixel. 
Using $i, j, k$ to index into row, column, class, 
\begin{align*}
\L_{\text{seg}} =\textstyle{\sum}_{i,j}\L_{\text{seg},i,j},\ \ 
\L_{\text{seg},i,j} & =\textstyle{\sum}_{k=0}^{K}(1-\hat{\p}_{i,j}^k)^{\gamma}\log(\hat{\p}_{i,j}^k) \y_{\ri,i,j}^k,
\end{align*}
where $\hat{\p}_{i,j}^k$ is the prediction logits of object class $k$
at pixel $(i,j)$, and $\gamma$ is the focusing parameter of focal loss \cite{lin2017focal}.

\subsubsection{Box Regression}

Two losses are involved in training the network to produce bounding boxes.

\paragraph{Heatmap Loss}

The heatmap loss is used to train the network to locate centers of the objects. Since we have separate detection heads for different classes, we construct the ground truth heatmap value $\y_{\hm, v}^k$ for each class $k$ at the Cartesian coordinates of each voxel $v$. 
\begin{align*}
\L_{\hm} =&\textstyle{\sum}_{k=1}^{K}\textstyle{\sum}_{v}\L_{\hm,v}^{k},\\
\L_{\hm,v}^{k}  =& (1-\hat{\h}_v^{k})^{\alpha}\log(\hat{\h}_v^{k})\mathbb{I}\{\y_{\hm,v}^{k}>1-\epsilon\} + \nonumber \\
 & (1-\y_{\hm,v}^{k})^{\beta}(\hat{\h}_v^{k})^{\alpha}\log(1-\hat{\h}_k^{k})\mathbb{I}\{\y_{\hm,v}^{k}\le1-\epsilon\},
\end{align*}
where $\hat{\h}_v^{k}$ denotes the predicted
heatmap value for class $k$ at voxel $v$, $\epsilon$ is added for numerical stability, $\alpha$ and $\beta$ are focusing hyper-parameters.

\paragraph{Shape Regression Loss}
For any $\y_{\hm, v}^k$ higher than a threshold (i.e. close to an object center), a bin loss \cite{shi20193d} is used to regress the object's heading, and the other box shape parameters are directly regressed under smooth L1 losses.
\section{Adaptation for Supervised Learning}
\label{sec: supervised}

In this section, we discuss how to use single-class labels to train a multi-class detector, which requires adaptation of the supervised learning -- more specifically adaptation of the detector training losses. In the next section, we will discuss how we can leverage pseudo labels to further improve the training effectiveness.

\subsection{Loss Modification}

\subsubsection{Foreground Point Selection}
\label{sec: seg3}

\begin{figure}
\centering{}\includegraphics[width=\linewidth]{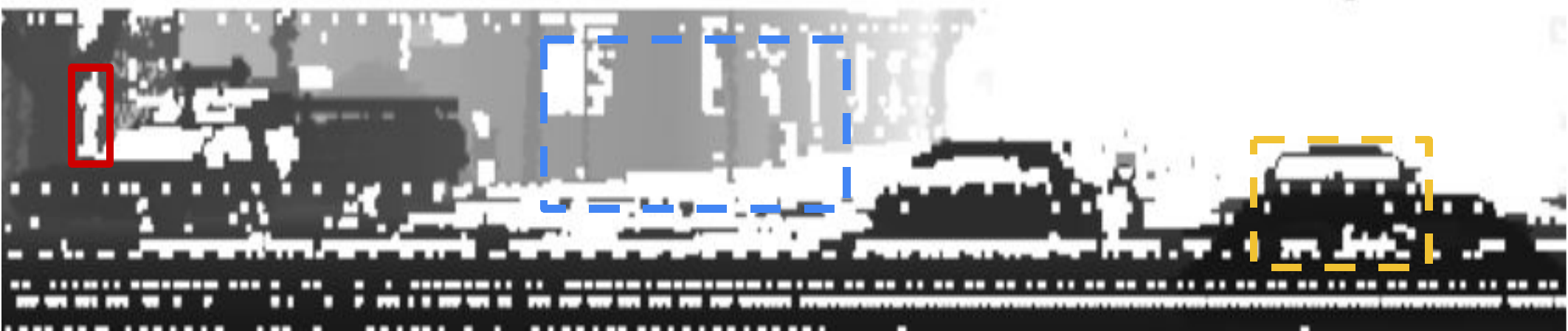}
\caption{A range image with only the pedestrian class labeled. Pixels labeled as pedestrian (red box) have accurate label. However, the other pixels have missing label: they may belong to background (blue dashed box) or vehicle (yellow dashed box) since the vehicle class is unlabeled.}
\label{fig: range}
\end{figure}

Consider a range image which contains both vehicles and pedestrians, but only pedestrians are labeled.
Although we can ensure that the pixels that are labeled as pedestrian have accurate labels, the pixels that are background-to-pedestrians might actually be pixels that belong to vehicles, rather
than the real background pixels (see Figure~\ref{fig: range} for illustration).
It is critical to handle those pixels with uncertain label properly.
We propose and study several strategies.

\paragraph{Aggressive Supervision}

Aggressive supervision is a simple approach that ignores the fact that pixels that are background-to-class-$k$ may not be the real background. It simply trains the model as if all the pixels were correctly labeled. This can still be practical, especially when the objects are sparse and thus most of the pixels that are labeled as background are truly background.
However, this solution essentially injects wrong information
into the training, which can be harmful.

\paragraph{Conservative Supervision}

An alternative to the Aggressive approach is a conservative
approach that does not produce loss on non-foreground pixels.
Specifically, given the segmentation label $\Y_{\ri}$, we define a set of pixels that have missing label and train the model using the modified segmentation loss 
\begin{align*}
\up & =\left\{ (i,j): \textstyle{\sum}_{k=1}^K \y_{\ri,i,j}^k=0 \right\}, 
\\
\L_{\text{seg}}^{\text{conservative}}& =\textstyle{\sum}_{i,j}\L_{\text{seg},i,j}\mathbb{I}\{(i,j)\notin\up\}.
\end{align*}
Although this conservative approach avoids injecting wrong information,  all the background pixels are masked out for training. As a result, none of the pixels will be classified as background, which would lead to a difficult second stage.

\paragraph{Informed Supervision}

Despite that the Aggressive and Conservative approaches are reasonable, they suffer from flaws such as injecting wrong supervision or lack of supervision on the background pixels. To improve, we propose a third approach that uses all the pixels for training and does not inject wrong supervision, which we name Informed supervision.

The idea is derived from the principal of maximum likelihood estimation. Let $\uc$ be the set of classes that are \emph{not} labeled in the current data.
For pixels in $\up$, although we do not know
its exact ground truth label, we know that they must not belong to
any classes that has ground truth label. Thus they can only
be of any class in $\uc$ or $0$ (background). 
For these pixels with missing label, we transform the original $K + 1$-way classification problem into a $K + 1 - |\uc|$-way classification problem by summing the prediction logits of classes in $\uc$ and $0$. The modified loss is as follows:

\begin{align*}
\L_{\seg,i,j}^{\text{informed}} & =\begin{cases}
\sum\limits_{k \notin \uc}(1-\hat{\p}_{i,j}^{\tilde{k}})^{\gamma}\log(\hat{\p}_{i,j}^{\tilde{k}})\y_{\ri,i,j}^{\tilde{k}} & \hspace{-1.5mm}\text{if}\ (i,j)\in\up\\
\L_{\seg,i,j} & \hspace{-1.5mm}\text{o/w}
\end{cases}
\end{align*}
where
\begin{align*}
 \hat{\p}_{i,j}^{\tilde{k}} & =
 \begin{cases}
 \sum_{c \in \{0\} \cup \uc} \hat{\p}_{i,j}^c & \text{if} \ k = 0 \\
 \hat{\p}_{i,j}^k & \text{o/w}
 \end{cases}
 \\
 \hspace{-2.5mm}
 \y_{\ri,i,j}^{\tilde{k}} & =
 \begin{cases}
 \mathbb{I}\{(i.j)\in \up\} & \text{if} \ k = 0 \\
 \y_{\ri,i,j}^k & \text{o/w}
 \end{cases}
\end{align*}

\subsubsection{Box Regression}
\label{sec: sup_box}

Adapting the box regression losses is straightforward. To train the
detection head of object class $k$, we only use the input data that labels class $k$:
\begin{equation*}
\L_{\hm}=\textstyle{\sum}_{k=1}^{K}\textstyle{\sum}_{v}\L_{\hm,v}^{k}\mathbb{I}\{k \notin \uc \}.
\end{equation*}
Note that the shape regression loss is automatically masked
out when there is no bounding boxes of the corresponding class, so no extra
modification is required for this loss.

\section{Adaptation for Learning with Pseudo Label}
\label{sec: pseudo}

In this section, we improve the multi-class detector performance by leveraging pseudo labels.

\subsection{Pseudo Label Generating Strategy}
Below we discuss several options to generate pseudo labels to augment the $K$ single-class datasets.

\subsubsection{Self Labeling}

The self labeling approach is adapted from FixMatch \cite{sohn2020fixmatch} for image classification, in which we generate the pseudo labels using the model itself. During training, we feed the data into the multi-class RSN and generate predicted classes for range image pixels and bounding boxes. Prediction with high confidence are saved as pseudo labels. Notice that we need to generate two realizations of data augmentation of the same input to prevent the model from degenerating into a trivial solution.

\subsubsection{Teacher Labeling}

Similarly, we can generate pseudo labels using $K$ well-trained teacher detectors, each being a standard RSN that detects a single class. The teacher model for class $k$ is trained using only the portion of the data that labels class $k$. We use the corresponding teacher model to generate pseudo labels for the unlabeled classes.

\subsubsection{Integrated Labeling}
We can also combine the two approaches above. For segmentation, we generate the pixel classes based on the ensemble predictions from the teacher and the trained model with equal weight. For bounding boxes, we combine the predicted boxes from all models and filter out overlapping boxes by Non-Maximum Suppression.

\subsection{Incorporating Pseudo Labels}

\begin{figure*}[t]
     \centering
     \begin{subfigure}[b]{0.33\textwidth}
         \centering
         \includegraphics[width=\linewidth]{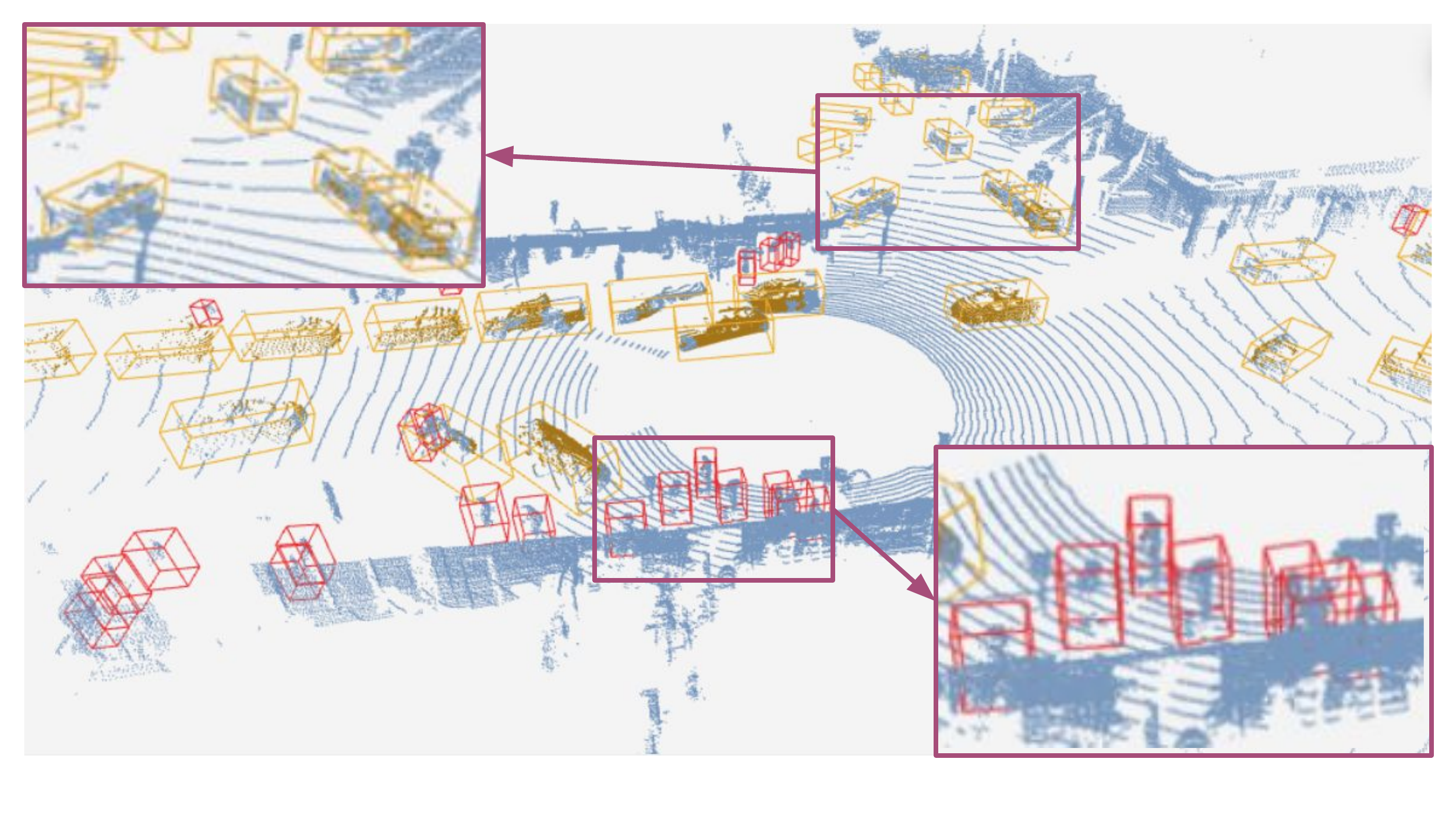}
         \vspace{-0.7cm}
         \caption{Aggressive Supervision}
         \label{fig: ri_abs}
     \end{subfigure}\hspace{-2mm}
     \begin{subfigure}[b]{0.33\textwidth}
         \centering
         \includegraphics[width=\linewidth]{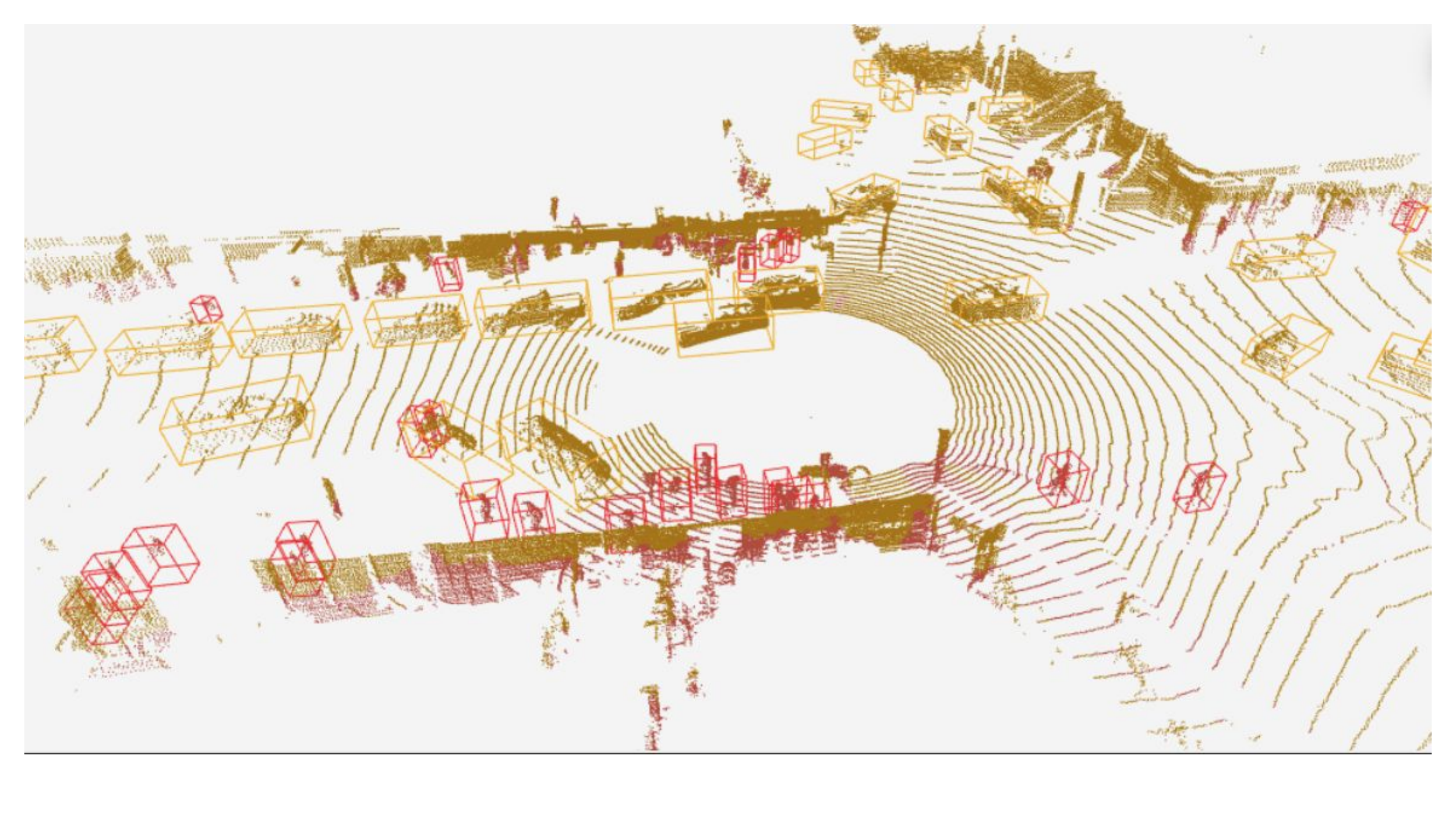}
         \vspace{-0.7cm}
         \caption{Conservation Supervision}
         \label{fig: ri_cs}
     \end{subfigure}\hspace{-2mm}
     \begin{subfigure}[b]{0.33\textwidth}
         \centering
         \includegraphics[width=\linewidth]{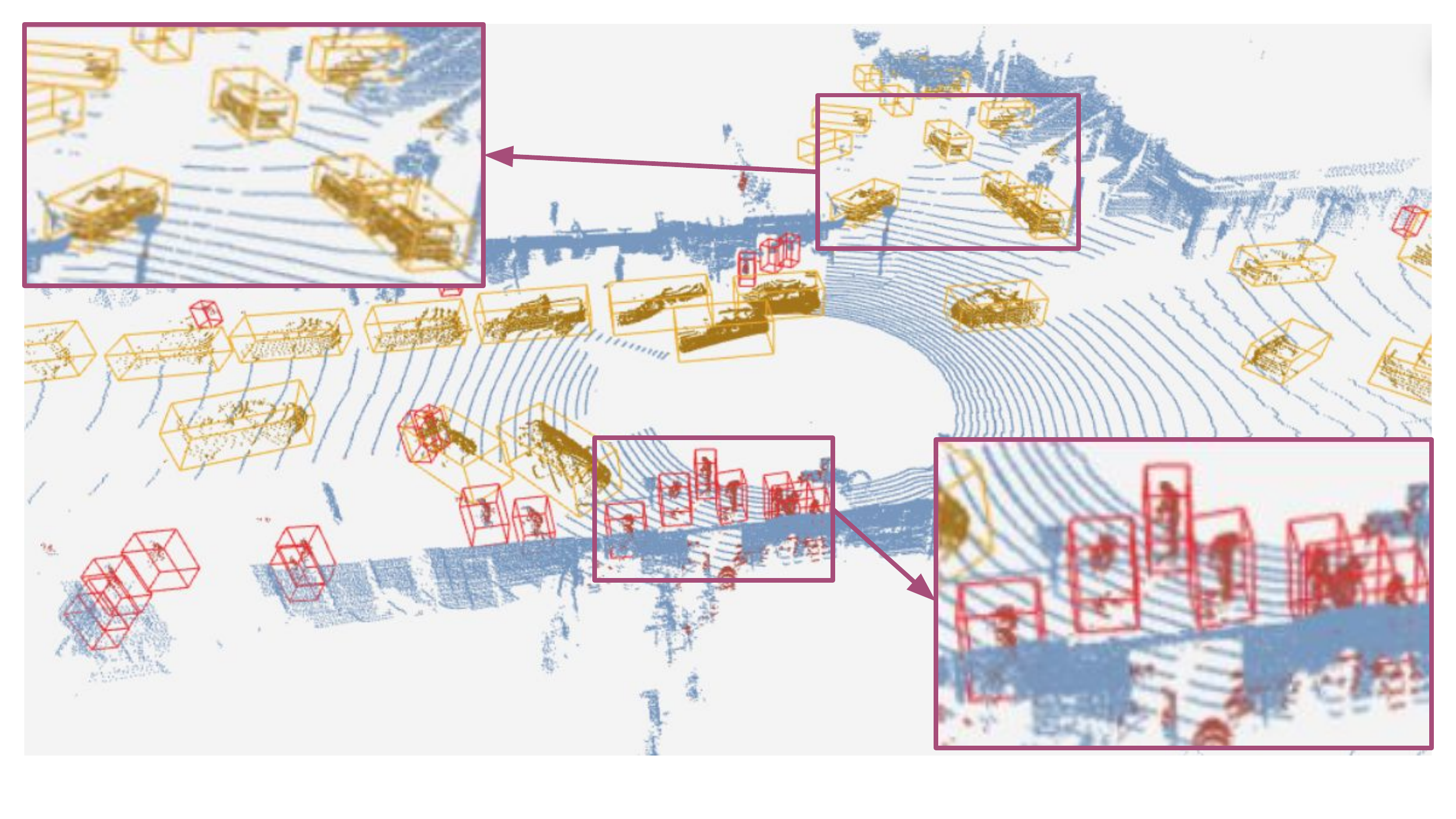}
         \vspace{-0.7cm}
         \caption{Informed Supervision}
         \label{fig: ri_is}
     \end{subfigure}
        \caption{Foreground point selection under different schemes. \emph{Boxes} are the ground truth. \emph{Points} being classified as background, vehicle and pedestrian are colored blue, yellow, and red. The Aggressive scheme misclassifies many object points as background. The Conservative scheme would prompt all points to be predicted as foreground. The Informed scheme gives accurate prediction, despite never having both vehicle and pedestrian labels on the same training data.}
        \label{fig: ri}
\end{figure*}

We follow \cite{sohn2020fixmatch} and only include pseudo labels
whose confidence scores are above a threshold. Specifically,
the final label for the range image is 
\[
\hat{\y}_{\ri,i,j}^{k}=\begin{cases}
\mathbb{I}\{ k = \arg\max_{c\in\{0\} \cup \uc}\hat{\p}_{i,j}^c \} & \text{if}\ (i,j)\in\up\\
\y_{\ri,i,j}^k & \text{o/w}
\end{cases}
\]
where $\hat{\p}_{i,j}^{c}$ is the prediction logits of the model
that generates the pseudo labels. Analogous to $\up$, we maintain a set $\cp$ to indicate pixels with ``non-trustworthy labels'', i.e., pixels that does not have ground truth label nor high confidence pseudo label
\[
\cp=\left\{ (i,j):\ \max_{c}\hat{\p}_{i,j}^{c}<\tau_{\text{pixel}}\ \text{and}\ (i,j)\in\up\right\} ,
\]
where $\tau_{\text{pixel}}$ is the threshold to decide whether the prediction is confident enough to be regarded as high quality.

Similarly, the final bounding box labels are the union of the original
ground truth boxes and the generated pseudo bounding boxes with scores higher than a threshold $\tau_{\bbox}$. 
\[
\hat{\Y}_{\bbox}^{k}=\begin{cases}
\{b:\score(b, k)\ge\tau_{\bbox}\} & \text{if}\ k \in \uc \\
\Y_{\bbox}^{k} & \text{o/w}
\end{cases}
\]
Here $\score(b, k)$ is the score of the pseudo bounding box of class $k$. Notice that the pseudo heatmap can be generated accordingly using the pseudo bounding boxes.

\subsection{Loss Modification}

\subsubsection{Foreground Point Selection}

Pseudo labels help reduce the amount of missing label pixels $\up$, but there are still pixels with non-trustworthy labels $\cp$.
Therefore, conveniently, the three schemes described in Section~\ref{sec: seg3} all still apply, simply by replacing $\up$ with $\cp$. Notice that under self labeling, the Conservative approach is still not suitable, as it will still classify all points as foreground.

\subsubsection{Box Regression}

With the pseudo labeled bounding boxes, the data that did not have $\uc$ originally labeled can now be utilized to train the detection
head of those classes. Given the pseudo labeled bounding boxes, similar
to supervised learning discussed in Section~\ref{sec: sup_box}, we do not need to modify the shape
regression other than simply replacing the ground truth box label by
the pseudo box label. 

For the heatmap loss, voxels that fall within any box that belongs to $\hat{\Y}_{\bbox}^{k}$ have trustworthy heatmap ground truth value.
However, we cannot ensure whether the remaining voxels are background or within a bounding box of class $u \in \uc$. Due to such uncertainty, the
heatmap loss needs special design and treatment. Similar to Section~\ref{sec: seg3}, we propose three schemes for modifying the heatmap loss.

\paragraph{Aggressive Supervision}

Aggressive supervision simply ignores the fact that some
of the background voxels may actually belong to an object, and trains the model as if the pseudo boxes were perfect. Its performance can be sensitive to the quality of the pseudo labels and may inject wrong information for training.

\paragraph{Conservative Supervision}

Following the same philosophy, Conservative supervision masks out losses on voxels without trustworthy labels. As a result, we only train on voxels with foreground labels. 
The heatmap loss becomes
\[
\hat{\L}_{\hm}^{\text{conservative}}=\textstyle{\sum}_{k=1}^{K}\textstyle{\sum}_{v}\hat{\L}_{\hm,v}^{k}\mathbb{I}\{v\in\uh^{k}\ \text{or}\ k\notin \uc \}
\]
Here $\uh^{k}=\{v:\exists b\in \hat{\Y}_{\bbox}^{k}\ \text{s.t.}\ v\ \text{is within }b\}$
denotes the set of voxels that fall within one of the trustworthy
bounding boxes, and $\hat{\L}_{\hm,v}^{k}$ is calculated based on $\hat{\Y}_{\bbox}^k$ as opposed to $\Y_{\bbox}^k$.

\paragraph{Informed Supervision}

The Informed supervision we derived for segmentation loss is not directly applicable here for heatmap loss. 
The main reason is that the segmentation is multi-class and mutual-exclusive (one out of $K$), while the heatmap loss is single-class ($k$ vs not $k$).

However the same spirit remains.
Here we utilize a special property of 3D detection: different from 2D detection where each pixel might belong to different objects (since objects can overlap with each other when projected onto a 2D image), in 3D point cloud, we can assume that each point only belongs to one class. 
That means for the detection head of class $k$, if a voxel belongs to any other foreground class, its heatmap value of class $k$ must be $0$ as this voxel cannot belong to any bounding boxes of class $k$. This gives the following modified heatmap loss:
\begin{align*}
 & \hat{\L}_{\hm}^{\text{informed}}=\textstyle{\sum}_{k=1}^{K}\textstyle{\sum}_{v}\hat{\L}_{\hm,v}^{k}\mathbb{I}\{v\in\uh\ \text{or}\ k\notin \uc \},\\
 & \uh=\{v:\exists b\in \cup_{k=1}^K \hat{\Y}_{\bbox}^{k} \ \text{s.t.}\ v\ \text{is within }b\}
\end{align*}
where, different from the Conservative supervision earlier, the construction of $\uh$ utilizes the information of (pseudo) bounding boxes of all classes (without superscript $k$).

\section{Experimental Results}

\begin{table*}
\centering{}%
\scalebox{0.94}{
\begin{tabular}{cc|cc|cc|cc|cc|cc|cc|cc|cc}
\toprule
\multirow{2}{*}{Algorithms} & \multirow{2}{*}{Scheme} & \multicolumn{4}{c|}{$\nicefrac{\text{\#V}}{\text{\#P}}=\unitfrac{90}{10}$} & \multicolumn{4}{c|}{$\nicefrac{\text{\#V}}{\text{\#P}}=\unitfrac{10}{90}$} & \multicolumn{4}{c|}{$\nicefrac{\text{\#V}}{\text{\#P}}=\unitfrac{95}{5}$} & \multicolumn{4}{c}{$\nicefrac{\text{\#V}}{\text{\#P}}=\unitfrac{5}{95}$}\tabularnewline
 &  & \multicolumn{2}{c}{Vehicle} & \multicolumn{2}{c|}{Pedestrian} & \multicolumn{2}{c}{Vehicle} & \multicolumn{2}{c|}{Pedestrian} & \multicolumn{2}{c}{Vehicle} & \multicolumn{2}{c|}{Pedestrian} & \multicolumn{2}{c}{Vehicle} & \multicolumn{2}{c}{Pedestrian}\tabularnewline
\hline 
Supervised & Aggressive & 70.2 & -1.8 & 53.5 & -18.9 & 63.5 & -8.5 & 68.3 & -4.1 & 69.8 & -2.2 & 28.9 & -43.5 & 18.4 & -53.6 & 69.2 & -3.2\tabularnewline
\hline 
Supervised & Informed & 70.2 & -1.8 & 59.3 & -13.1 & 65.4 & -6.6 & 71.6 & -0.8 & 70.7 & -1.3 & 46.0 & -26.4 & 61.8 & -10.2 & 72.5 & +0.1\tabularnewline
Self Label & Informed & 70.2 & -1.8 & 62.1 & -10.3 & 66.0 & -6.4 & 72.0 & -0.4 & 70.9 & -1.1 & 47.3 & -25.1 & 64.1 & -7.9 & 72.3 & -0.1\tabularnewline
Teacher Label & Informed & \pmb{71.7} & \pmb{-0.3} & 67.1 & -5.3 & 68.8 & -3.6 & 73.2 & +0.8 & 71.3 & -0.7 & 57.8 & -14.6 & 66.0 & -6.0 & 71.7 & -0.8\tabularnewline
Integrated Label & Informed & 71.6 & -0.4 & \pmb{68.5} & \pmb{-3.9} & \pmb{69.0} & \pmb{-3.0} & \pmb{73.8} & \pmb{+1.4} & \pmb{71.5} & \pmb{-0.5} & \pmb{59.5} & \pmb{-12.9} & \pmb{66.3} & \pmb{-5.7} & \pmb{73.2} & \pmb{+0.8}\tabularnewline
\hline 
\gray{Full Label} & \gray{-} & \gray{72.0} & \gray{0.0} & \gray{72.4} & \gray{0.0} & \gray{72.0} & \gray{0.0} & \gray{72.4} & \gray{0.0} & \gray{72.0} & \gray{0.0} & \gray{72.4} & \gray{0.0} & \gray{72.0} & \gray{0.0} & \gray{72.4} & \gray{0.0}\tabularnewline
\bottomrule
\end{tabular}
}
\caption{Comparing algorithms and schemes developed for single-class supervision (SCS). The two adjacent numbers are the detection AP and its gap to the full label upper bound. Notice that the last row uses more labels and does \emph{not} belong to SCS, hence marked in \gray{gray} to indicate that these numbers are unfair comparisons.} \label{tbl: algos}
\vspace{-0.2cm}
\end{table*}

We perform experiments on the Waymo Open Dataset (WOD) \cite{sun2020scalability}, specifically detecting vehicles and pedestrians (i.e., $K= 2$).
WOD provides 798 training sequences with all $K$ classes labeled, so we create our single-class supervision scenario by first dividing these sequences into two disjoint sets, and then masking out labels for either class on the corresponding set.
Therefore, our experiments correspond to roughly $50\%$ labeling cost savings.
For the division of sequences, we consider two challenging and imbalance settings 10\%-90\% and 5\%-95\%. We denote the setting that $x\%$, $(100-x)\%$ sequences contains only vehicle, pedestrian labels as $\nicefrac{\text{\#V}}{\text{\#P}}=\unitfrac{x}{100-x}$.
The evaluation on the validation set remains unchanged from the standard setting.

We employ the same training protocol for all experiments, tuned based on the adapted supervised learning.
We train with batch size 64 for 80K iterations with 4K warmup. 
The number of channels in each layer of our multi-class RSN is 3/4 that of CarL\_1f / PedL\_1f in \cite{sun2021rsn}, to reduce the memory cost.
The remaining hyper-parameters and data augmentation (random flipping and rotation are applied) are the same as those in \cite{sun2021rsn}. For self labeling and integrated labeling, we do not apply augmentation when generating pseudo labels using the multi-class RSN.
Since the sizes of the two subsets are quite different, we also apply dataset resampling such that each data in the mini-batch is drawn from the two subsets with equal probability. 

\subsection{Algorithm Comparison} \label{sec: algo compare}
We compare different methods to learn the multi-class detector from single-class labels in Table~\ref{tbl: algos}.
The evaluation metric is 3D AP with L1 difficulty.

We consider the supervised learning algorithm with Aggressive scheme to be our \emph{baseline} (as we are in a new problem setting, there is no prior baseline from other works), and consider the performance of our multi-class RSN when trained on the unmasked, fully labeled data (which does not belong to SCS) to be the performance \emph{upper bound}.

Within supervised learning, Informed supervision outperforms the Aggressive supervision baseline, showing the superiority of proper SCS modeling.
As is evident from Figure~\ref{fig: ri}, Aggressive supervision misclassifies many object points as background and by the design of RSN, those points will not be included in proposing bounding boxes.
By comparison, Informed supervision gives accurate prediction for all three kinds of points in Figure~\ref{fig: range}, even though there are no points explicitly labeled as background. 
We do not include quantitative results with Conservative supervision as all points are classified as foreground, causing failed training of the second stage in RSN, though we can still visualize its first stage in Figure~\ref{fig: ri}.

We then use the Informed scheme to compare between supervised learning and pseudo labeling. 
We observe that the adapted pseudo labeling can give significant improvement over adapted supervised learning. 
Among the pseudo labeling algorithms, the integrated labeling delivers the best performance, as expected.
The teacher labeling outperforms the self labeling approach, where the teacher models are single-class detectors trained only on the subset of sequences labeled with the corresponding class.
One of the reasons may be that the different object features required by vehicle and pedestrian cause conflicts in the hidden representation given the limited model capacity, and thus the quality of the pseudo labels generated by the two \emph{single}-class detectors are better than those generated by the \emph{multi}-class detector. 

The fact that detecting vehicle and pedestrian may be conflicting may also explain the occasional cases where the pseudo labeling algorithm surpasses the full label upper bound in detecting one class (e.g. in $\nicefrac{\text{\#V}}{\text{\#P}}=\unitfrac{90}{10}$, teacher and integrated labeling outperform the full label upper bound in detecting pedestrian). In cases where both classes have sufficient labels, the two tasks compete with each other and the model learns a representation that performs well for both tasks. While, in these cases, when there is not enough information for the model to learn a strong representation for vehicle, the representation learned by the model will favor detecting the pedestrian.

\subsection{Ablations on Modeling Missing Label} \label{sec: ablation missing}

Under the teacher labeling algorithm ($\nicefrac{\text{\#V}}{\text{\#P}}=\unitfrac{90}{10}$), we ablate the effects of the three schemes we proposed (Aggressive, Conservative, Informed) on either the segmentation loss or the heatmap loss.
The results are summarized in Table~\ref{tbl: ablation missing label}.
Employing Informed supervision for both segmentation and heatmap results in the best performance, as Informed supervision best exploits trustworthy label information. Aggressive supervision is worse than the Conservative approach. This holds for both the segmentation loss and the heatmap loss. We believe this shows the negative influence of injecting \emph{wrong} information outweighs the benefit of injecting \emph{more} information.

\begin{table}
\centering{}%
\scalebox{0.94}{
\begin{tabular}{cc|cc|cc}
\toprule
Segmentation & Heatmap & \multicolumn{2}{c}{Vehicle} & \multicolumn{2}{c}{Pedestrian}\tabularnewline
\hline 
Aggressive & Informed & 69.3 & -3.4 & 54.1 & -17.0\tabularnewline
Conservative & Informed & 70.8 & -0.9 & 65.6 & -1.5\tabularnewline
Informed & Aggressive & 65.6 & -6.1 & 48.8 & -18.3\tabularnewline
Informed & Conservative & 71.0 & -0.7 & 66.3 & -0.8\tabularnewline
\hline 
Informed & Informed & 71.7 & 0.0 & 67.1 & 0.0\tabularnewline
\bottomrule
\end{tabular}
}
\caption{Comparing different schemes for handling segmentation / heatmap loss with missing label. The two adjacent numbers are the detection AP and its performance gap to employing Informed supervision for both.}
\vspace{-0.2cm}
\label{tbl: ablation missing label}
\end{table}

\subsection{Dataset Resampling} \label{sec: resample}
\begin{figure}
\centering{}
\includegraphics[width=\linewidth]{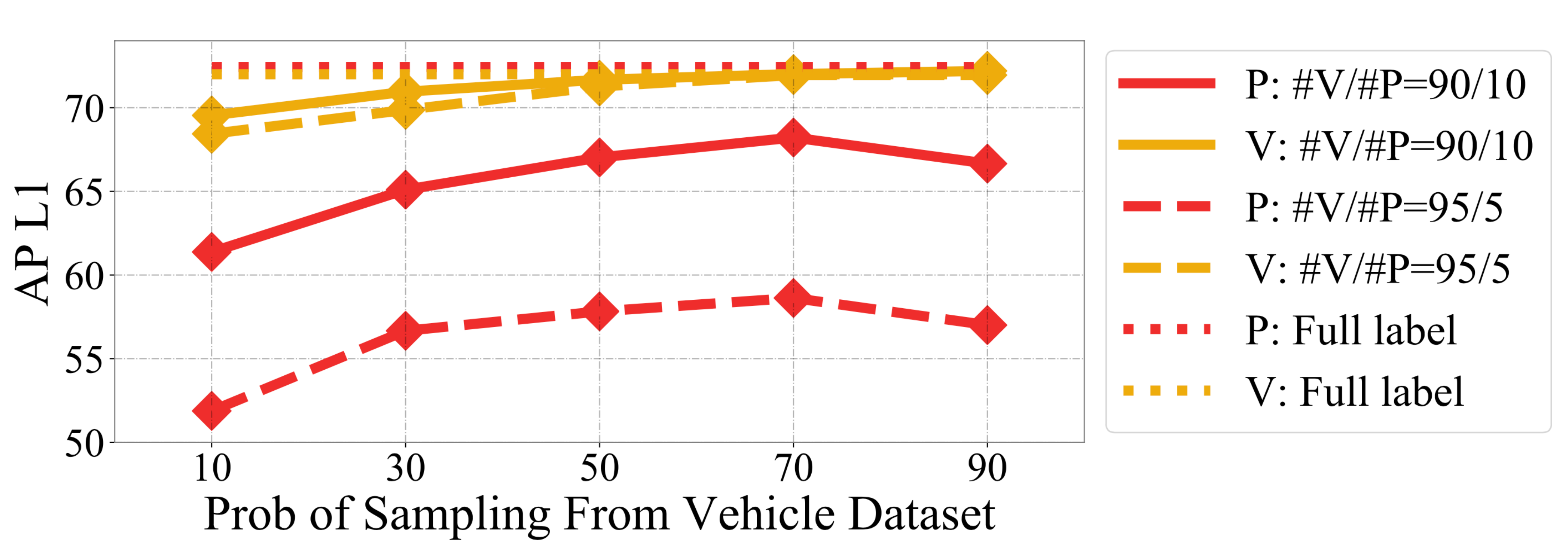}
\caption{Detection AP with varying dataset resampling probability. P / V in the legend stand for Pedestrian / Vehicle.} \label{fig: resample}
\vspace{-0.2cm}
\end{figure}

After analyzing the Table~\ref{tbl: algos} vertically, we now analyze it horizontally.
The general trend is that when a class has fewer data, the AP on this class is lower and has more room to improve, which is expected. However, there are a few occasions where having more data of a class results in worse AP, for example detecting vehicles using teacher labeling at $\nicefrac{\text{\#V}}{\text{\#P}}=\unitfrac{90}{10}$ vs $\nicefrac{\text{\#V}}{\text{\#P}}=\unitfrac{95}{5}$ ($71.7$ vs $71.3$). We found the cause to be dataset resampling.
In Figure~\ref{fig: resample} we vary the probability of sampling images from the dominant vehicle class when $\nicefrac{\text{\#V}}{\text{\#P}}=\unitfrac{90}{10}$ or $\nicefrac{\text{\#V}}{\text{\#P}}=\unitfrac{95}{5}$ (default is 50\%, i.e., equal probability).
Towards the right side of the figure, when we sample predominantly from the vehicle class when the amount of labeled vehicle sequences is already dominant, the performance of the minority pedestrian class worsens drastically. 
On the other hand, towards the left side, when we sample predominantly from the minority pedestrian class, the model sees too many repetitive pedestrians and not enough vehicle epochs, degrading the performance on both tasks. Therefore, it is critical to find the middle sweet spot. 

\section{Conclusion}

We study training a multi-class 3D object detector under the ``single-class supervision'' learning setting, by proposing and benchmarking various baselines and strategies.
Notably, our proposed Informed Supervision combined with pseudo labeling can approach or match the upper bound that is full supervision, saving significant labeling costs. 

For future work, we plan to expand the applications of SCS.
Within 3D object detection, we plan to experiment with more classes and more diverse architectures.
Beyond 3D object detection, we plan to expand to other tasks such as 2D object detection, and move from multi-class to multi-task (e.g. joint detection and segmentation).

\bibliography{root}
\bibliographystyle{IEEEtran}

\addtolength{\textheight}{-12cm}   





\end{document}